# An Investigation of the Sampling-based Alignment Method and Its Contributions

Juan Luo[1] and Yves Lepage[2]

[1,2]Graduate School of Information, Production and Systems, Waseda University
2-7 Hibikino, Wakamatsu-ku, Fukuoka 808-0135, Japan
[1]`juan.luo@suou.waseda.jp` and [2]`yves.lepage@waseda.jp`

## ABSTRACT

*By investigating the distribution of phrase pairs in phrase translation tables, the work in this paper describes an approach to increase the number of n-gram alignments in phrase translation tables output by a sampling-based alignment method. This approach consists in enforcing the alignment of n-grams in distinct translation subtables so as to increase the number of n-grams. Standard normal distribution is used to allot alignment time among translation subtables, which results in adjustment of the distribution of n-grams. This leads to better evaluation results on statistical machine translation tasks than the original sampling-based alignment approach. Furthermore, the translation quality obtained by merging phrase translation tables computed from the sampling-based alignment method and from MGIZA++ is examined.*

## KEYWORDS

*Alignment, Phrase Translation Table, Statistical Machine Translation Task.*

## 1. INTRODUCTION

Sub-sentential alignment plays an important role in the process of building a machine translation system. The quality of the sub-sentential alignment, which identifies the relations between words or phrases in the source language and those in the target language, is crucial for the final results and the quality of a machine translation system. Currently, the most widely used state-of-the-art alignment tool is GIZA++ [1], which belongs to the estimating trend. It trains the ubiquitous IBM models [2] and the HMM introduced by [3]. MGIZA++ is a multi-threaded word aligner based on GIZA++, originally proposed by [4].

In this paper, we focus on investigating a different alignment approach to the production of phrase translation tables: the sampling-based approach [5]. There are two contributions of this paper:

- Firstly, we propose a method to improve the performance of this sampling-based alignment approach;
- Secondly, although evaluation results show that it lags behind MGIZA++, we show that, in combination with the state-of-the-art method, it slightly outperforms MGIZA++ alone and helps significantly to reduce the number of out-of-vocabulary words.

The paper is organized as follows. Section 2 presents related work. In Section 3, we briefly review the technique of sampling-based alignment method. In Section 4, we propose a variant in order to improve its performance. We also introduce standard normal distribution of time to bias

 9

the distribution of n-grams in phrase translation tables. Section 5 presents results obtained by merging two aligners' phrase translation tables. Finally, in Section 6, conclusions and possible directions for future work are presented.

## 2. RELATED WORK

There are various methods and models being suggested and implemented to solve the problem of alignment. One can identify two trends to solve this problem [6]. On one side, there is the associative alignment trend, which is illustrated by [7, 8, 9]. On the other side, the estimating trend is illustrated by [1, 2, 10].

Associative alignment method employs similarity measures and association tests. These measures are meant to rank and determine if word pairs are strongly associated with each other. In [7], Gale and Church propose to use measures of association to find correspondences between words. They introduce the $\Phi^2$ coefficient, based on a two by two contingency table. Melamed [8] shows that most source words tend to correspond to only one target word and presented methods for biasing statistical translation models, which leads to positive impact on identifying translational equivalence. In [9], Moore proposes the log-likelihood-ratio association measure and alignment algorithm, which is faster and simpler than the generative probabilistic framework. The estimating alignment approach employs statistical models and the parameters are estimated through maximization process. In [1, 2], a set of word alignment models are introduced and phrase alignments are extracted given these word alignments. Liang et al. [10] propose a symmetric alignment, which trains two asymmetric models jointly to maximize agreement between the models.

## 3. SAMPLING-BASED ALIGNMENT METHOD

The sampling-based approach is implemented in a free open-source tool called Anymalign (http://anymalign.limsi.fr/). It is in line with the associative alignment trend and it is much simpler than the models implemented in MGIZA++. The sampling-based alignment approach takes as input a sentence-aligned corpus and output pairs of sub-sentential sequences similar to those in phrase translation tables, in a single step. The approach exploits low frequency terms and relies on distribution similarities to extract sub-sentential alignments. In addition, it has been shown in [11] that the sampling-based method, i.e., Anymalign, requires less memory in comparison with GIZA++. As a last and remarkable feature, it is capable of aligning multiple languages simultaneously [5], but we will not use this feature in this paper as we will restrain ourselves to bilingual experiments.

In sampling-based alignment, low frequency terms and distribution similarities lay the foundation for sub-sentential alignment. Low frequency terms, especially *hapaxes*, have been shown to safely align across languages [12]. Hapaxes are words that occur only once in the input corpus. It has been observed that the translational equivalence between hapaxes, which co-occur together in a parallel sentence, is highly reliable. Aligned hapaxes have exactly the same trivial distribution on lines (Here, "line" denotes a (source, target) sentence pair in a parallel corpus): 0 everywhere, except 1 on the unique line they appear in. On the other end of the frequency spectrum, fullstops at the end of each sentence in both source and target languages have the same trivial distribution on lines if one line contains one sentence: 1 everywhere. Building on these observations and, exploiting the possibility of sampling a corpus in many sub corpora, only those sequences of words sharing the exact same distribution (i.e., they appear exactly in the same sentences of the corpus) are considered for alignment.

The key idea is to make more words share the same distribution by artificially reducing their frequency in multiple random subcorpora obtained by sampling. The distribution here is denoted as the co-occurrences between words in the context of parallel sentences. Indeed, the smaller a subcorpus, the less frequent its words, and the more likely they are to share the same distribution; hence the higher the proportion of words aligned in this subcorpus.

The subcorpus selection process is guided by a probability distribution which ensures a proper coverage of the input parallel corpus:

$$p(k) = \frac{-1}{k \log(1 - k / n)} \text{ (to be normalized)} \qquad (1)$$

where $k$ denotes the size (number of sentences) of a subcorpus and $n$ the size of the complete input corpus. Note that this function is very close to $1/k^2$: it gives much more credit to small subcorpora, which happen to be the most productive [5]. Once the size of a subcorpus has been chosen according to this distribution, its sentences are randomly selected from the complete input corpus according to a uniform distribution. Then, from each subcorpus, sequences of words that share the same distribution are extracted to constitute alignments along with the number of times they were aligned (contrary to the widely used terminology where it denotes a set of links between the source and target words of a sentence pair, we call "alignment" a (source, target) phrase pair, i.e., it corresponds to an entry in the so-called phrase translation tables). Eventually, the list of alignments is turned into a full-fledged phrase translation table, by calculating various features for each alignment. In the following, we use two translation probabilities and two lexical weights as proposed by [13], as well as the commonly used phrase penalty, for a total of five features.

One important characteristic of the sampling-based alignment method is that it is implemented with an *anytime* algorithm: the number of random subcorpora to be processed is not set in advance, so the alignment process can be interrupted at any moment. Contrary to many approaches, after a very short amount of time, *quality* is no more a matter of time, however *quantity* is: the longer the aligner runs (i.e. the more subcorpora processed), the more alignments produced, and the more reliable their associated translation probabilities, as they are calculated on the basis of the number of time each alignment was obtained. This is possible because high frequency alignments are quickly output with a fairly good estimation of their translation probabilities. As time goes, their estimation is refined, while less frequent alignments are output in addition.

Intuitively, since the sampling-based alignment process can be interrupted without sacrificing the quality of alignments, it should be possible to allot more processing time for n-grams of similar lengths in both languages and less time to very different lengths. For instance, a source bigram is much less likely to be aligned with a target 9-gram than with a bigram or a trigram. The experiments reported in this paper make use of the anytime feature of Anymalign and of the possibility of allotting time freely.

## 3.1. Preliminary Experiment

In order to measure the performance of the sampling-based alignment approach implemented in Anymalign in statistical machine translation tasks, we conducted a preliminary experiment and compared with the standard alignment setting: symmetric alignments obtained from MGIZA++. Although Anymalign and MGIZA++ are both capable of parallel processing, for fair comparison in time, we run them as single processes in all our experiments.

### 3.1.1. Experimental Setup

A sample of the French-English parts of the Europarl parallel corpus [14] was used for training, tuning and testing. A detailed description of the data used in the experiments is given in Table 1. The training corpus is made of 100k sentences. The development set contains 500 sentences, and 1,000 sentences were used for testing. To perform the experiments, a standard statistical machine translation system was built for each different alignment setting, using the Moses decoder [15], MERT (Minimum Error Rate Training) to tune the parameters of translation tables [16], and the SRI Language Modeling toolkit [17] to build the target language model. As for the evaluation of translations, four standard automatic evaluation metrics were used: WER [18], BLEU [19], NIST [20], and TER [21].

Table 1. Statistics on the French-English parallel corpus used for the training, development, and test sets.

|  |  | French | English |
|---|---|---|---|
| Train | sentences | 100,000 | 100,000 |
|  | words | 3,986,438 | 2,824,579 |
|  | words/sentence | 38 | 27 |
| Dev | sentences | 500 | 500 |
|  | words | 18,120 | 13,261 |
|  | words/sentence | 36 | 26 |
| Test | sentences | 1,000 | 1,000 |
|  | words | 38,936 | 27,965 |
|  | words/sentence | 37 | 27 |

### 3.1.2. Problem Definition

In a first setting, we evaluated the quality of translations output by the Moses decoder using the phrase translation table obtained by making MGIZA++'s alignments symmetric. In a second setting, this phrase translation table was simply replaced by that produced by Anymalign. Since Anymalign can be stopped at any time, for a fair comparison, it was run for the same amount of time as MGIZA++: seven hours in total. The experimental results are shown in Table 2.

Table 2. Evaluation results on a statistical machine translation task using phrase tables obtained from MGIZA++ and Anymalign (baseline).

|  | BLEU | NIST | WER | TER |
|---|---|---|---|---|
| MGIZA++ | **0.2742** | **6.6747** | **0.5714** | **0.6170** |
| Anymalign | 0.2285 | 6.0764 | 0.6186 | 0.6634 |

In order to investigate the differences between MGIZA++ and Anymalign phrase translation tables, we analyzed the distribution of n-grams of both aligners. The distributions are shown in Table 6(a) and Table 6(b). In Anymalign's phrase translation table, the number of alignments is 8 times that of 1×1 n-grams in MGIZA++ translation table, or twice the number of 1×2 n-grams or 2×1 n-grams in MGIZA++ translation table. Along the diagonal (m×m n-grams), the number of alignments in Anymalign table is more than 10 times less than in MGIZA++ table. This confirms the results given in [22] that the sampling-based approach excels in aligning unigrams, which makes it better at multilingual lexicon induction than, e.g., MGIZA++. However, its phrase translation tables do not reach the performance of symmetric alignments from MGIZA++ on translation tasks. This basically comes from the fact that Anymalign does not align enough long n-grams [22]. Longer n-grams are essential in a phrase-based machine translation system as they contribute to the fluency of translations.

# 4. DIVIDING INTO PHRASE TRANSLATION SUBTABLES

## 4.1. Enforcing Alignment of N-grams

To solve the above-mentioned problem, we propose a method to force the sampling-based approach to align more n-grams.

Consider that we have a parallel input corpus, i.e., a list of (source, target) sentence pairs, for instance, in French and English. Groups of characters that are separated by spaces in these sentences are considered as words. Single words are referred to as unigrams, and sequences of two and three words are called bigrams and trigrams, respectively. Theoretically, since the sampling-based alignment method excels at aligning unigrams, we could improve it by making it align bigrams, trigrams, or even longer n-grams as if they were unigrams. We do this by replacing spaces between words by underscore symbols and reduplicating words as many times as needed, which allows making bigrams, trigrams, and longer n-grams appear as unigrams. Table 3 depicts the way of forcing n-grams into unigrams.

Table 3. Transforming n-grams into unigrams by inserting underscores and reduplicating words for both the French part and English part of the input parallel corpus.

| n | French | English |
|---|--------|---------|
| 1 | le debat est clos . | the debate is closed . |
| 2 | le_debat  debat_est  est_clos  clos_. | the_debate  debate_is  is_closed  closed_. |
| 3 | le_debat_est  debat_est_clos  est_clos_. | the_debate_is debate_is_closed   is_closed_. |
| 4 | le_debat_est_clos  debat_est_clos_. | the_debate_is_closed  debate_is_closed_. |
| 5 | le_debat_est_clos_. | the_debate_is_closed_. |

Similar works on the idea of enlarging n-grams have been reported in [23], in which "word packing" is used to obtain 1-to-$n$ alignments based on co-occurrence frequencies, and [24], in which collocation segmentation is performed on bilingual corpus to extract $n$-to-$m$ alignments.

## 4.2. Phrase Translation Subtables

It is thus possible to use various parallel corpora, with different segmentation schemes in the source and target parts. We refer to a parallel corpus where source n-grams and target m-grams are assimilated to unigrams as an *unigramized n-m corpus*. These corpora are then used as input to Anymalign to produce phrase translation subtables, as shown in Table 4. Practically, we call Anymalign1-N the process of running Anymalign with all possible unigramized $n$-$m$ corpora, with $n$ and $m$ both ranging from 1 to a given N. In total, Anymalign is thus run N×N times. All phrase translation subtables are finally merged together into one large translation table, where translation probabilities are re-estimated given the complete set of alignments.

Table 4. List of n-gram translation subtables (TT) generated from the training corpus. These subtables are then merged together into a single phrase translation table.

| | | Target | | | | |
|---|---|---|---|---|---|---|
| | | 1-grams | 2-grams | 3-grams | … | N-grams |
| Source | 1-grams | TT1×1 | TT1×2 | TT1×3 | … | TT1×N |
| | 2-grams | TT2×1 | TT2×2 | TT2×3 | … | TT2×N |
| | 3-grams | TT3×1 | TT3×2 | TT3×3 | … | TT3×N |
| | … | … | … | … | … | … |
| | N-grams | TTN×1 | TTN×2 | TTN×3 | … | TTN×N |

Although Anymalign is capable of directly producing alignments of sequences of words, we use it with a simple filter (option -N 1 in the program), so that it only produces (typographic) unigrams in output, i.e., n-grams and m-grams assimilated to unigrams in the input corpus. This choice was made because it is useless to produce alignment of sequences of words, since we are only interested in *phrases* in the subsequent machine translation tasks. Those phrases are already contained in our (typographic) unigrams: all we need to do to get the original segmentation is to remove underscores from the alignments.

## 4.3. Equal Time Configuration

The same experimental process (i.e., replacing the translation table), as in the preliminary experiment, was carried out on Anymalign1-N with equal time distribution, which is, uniformly distributed time among subtables. For a fair comparison, the same amount of time was given: seven hours in total. The results are shown in Table 7. On the whole, MGIZA++ significantly outperforms Anymalign, by more than 4 BLEU points. The proposed approach (Anymalign1-N) produces better results than Anymalign in its basic version, with the best increase with Anymalign1-3 or Anymalign1-4 (+1.3 BP).

The comparison of Table 6(a) and Table 6(c) shows that Anymalign1-N delivers too many alignments outside of the diagonal (*m×m* n-grams) and still not enough along the diagonal. Consequently, this number of alignments should be lowered. A way of doing so is by giving less time for alignments outside of the diagonal.

## 4.4. Standard Normal Time Distribution

In order to increase the number of phrase pairs along the diagonal of the translation table matrix and decrease this number outside the diagonal (Table 4), we distribute the total alignment time among translation subtables according to the standard normal distribution as it is the most natural distribution intuitively fitting the distribution observed in Table 6(a).

$$\phi(n,m) = \frac{1}{\sqrt{2\pi}} e^{-\frac{1}{2}(n-m)^2} \qquad (2)$$

The alignment time allotted to the subtable between source *n*-grams and target *m*-grams will thus be proportional to φ(*n,m*). Table 5 shows an example of alignment times allotted to each subtable up to 4-grams, for a total processing time of 7 hours.

Table 5. Alignment time in seconds allotted to each unigramized parallel corpus of Anymalign1-4. The sum of the figures in all cells amounts to seven hours (7 hrs = 25,200 seconds).

|  |  | Target | | | |
|---|---|---|---|---|---|
|  |  | 1-grams | 2-grams | 3-grams | 4-grams |
| Source | 1-grams | 3,072 | 1,863 | 416 | 34 |
|  | 2-grams | 1,863 | 3,072 | 1,863 | 416 |
|  | 3-grams | 416 | 1,863 | 3,072 | 1,863 |
|  | 4-grams | 34 | 416 | 1,863 | 3,072 |

We performed a third evaluation using the standard normal distribution of time, as in previous experiments, again with a total amount of processing time (7 hours).

The comparison between MGIZA++, Anymalign in its standard use (baseline), and Anymalign1-N with standard normal time distribution is shown in Table 7. Anymalign1-4 shows the best performance in terms of BLEU and WER scores, while Anymalign1-3 gets the best results for the two other evaluation metrics. There is an increase in BLEU scores for almost all Anymalign1-N, from Anymalign1-3 to Anymalign1-10, when compared with the translation qualities of Anymalign1-N with equal time configuration (Table 7). The greatest increase in BLEU is obtained for Anymalign1-10 (almost +2 BP). Anymalign1-4 shows the best translation qualities among all other settings, but gets a less significant improvement (+0.2 BP).

Table 6. Distribution of phrase pairs in phrase translation tables.

(a) MGIZA++

| | | target | | | | | | |
|---|---|---|---|---|---|---|---|---|
| | | 1-grams | 2-grams | 3-grams | 4-grams | 5-grams | 6-grams | 7-grams | total |
| source | 1-grams | **89,788** | 44,941 | 10,700 | 2,388 | 486 | 133 | 52 | 148,488 |
| | 2-grams | 61,007 | **288,394** | 86,978 | 20,372 | 5,142 | 1,163 | 344 | 463,400 |
| | 3-grams | 19,235 | 149,971 | **373,991** | 105,449 | 27,534 | 7,414 | 1,857 | 685,451 |
| | 4-grams | 5,070 | 47,848 | 193,677 | **335,837** | 106,467 | 31,011 | 9,261 | 729,171 |
| | 5-grams | 1,209 | 13,984 | 73,068 | 193,260 | 270,615 | 98,895 | 32,349 | 683,380 |
| | 6-grams | 332 | 3,856 | 24,333 | 87,244 | 177,554 | 214,189 | 88,700 | 596,208 |
| | 7-grams | 113 | 1,103 | 7,768 | 33,278 | 91,355 | 157,653 | 171,049 | 462,319 |
| | total | 176,754 | 550,097 | 770,515 | 777,828 | 679,153 | 510,458 | 303,612 | 3,768,417 |

(b) Anymalign (baseline)

| | | target | | | | | | | |
|---|---|---|---|---|---|---|---|---|---|
| | | 1-grams | 2-grams | 3-grams | 4-grams | 5-grams | 6-grams | 7-grams | … | total |
| source | 1-grams | **791,099** | 105,961 | 9,139 | 1,125 | 233 | 72 | 37 | … | 1,012,473 |
| | 2-grams | 104,633 | **21,602** | 4,035 | 919 | 290 | 100 | 44 | … | 226,176 |
| | 3-grams | 10,665 | 4,361 | **2,570** | 1,163 | 553 | 240 | 96 | … | 92,268 |
| | 4-grams | 1,698 | 1,309 | 1,492 | **1,782** | 1,158 | 573 | 267 | … | 61,562 |
| | 5-grams | 378 | 526 | 905 | 1,476 | 1,732 | 1,206 | 642 | … | 47,139 |
| | 6-grams | 110 | 226 | 467 | 958 | 1,559 | 1,694 | 1,245 | … | 40,174 |
| | 7-grams | 40 | 86 | 238 | 536 | 1,054 | 1,588 | 1,666 | … | 35,753 |
| | … | … | … | … | … | … | … | … | … | … |
| | total | 1,022,594 | 230,400 | 86,830 | 55,534 | 42,891 | 37,246 | 34,531 | … | 1,371,865 |

(c) Anymalign1-4 (equal time configuration)

| | | target | | | | | | |
|---|---|---|---|---|---|---|---|---|
| | | 1-grams | 2-grams | 3-grams | 4-grams | 5-grams | 6-grams | 7-grams | total |
| source | 1-grams | **171,077** | 118,848 | 39,253 | 13,327 | 0 | 0 | 0 | 342,505 |
| | 2-grams | 119,953 | **142,721** | 67,872 | 24,908 | 0 | 0 | 0 | 355,454 |
| | 3-grams | 45,154 | 75,607 | **86,181** | 42,748 | 0 | 0 | 0 | 249,690 |
| | 4-grams | 15,514 | 30,146 | 54,017 | **60,101** | 0 | 0 | 0 | 159,778 |
| | 5-grams | 0 | 0 | 0 | 0 | 0 | 0 | 0 | 0 |
| | 6-grams | 0 | 0 | 0 | 0 | 0 | 0 | 0 | 0 |
| | 7-grams | 0 | 0 | 0 | 0 | 0 | 0 | 0 | 0 |
| | total | 351,698 | 367,322 | 247,323 | 141,084 | 0 | 0 | 0 | 1,107,427 |

(d) Anymalign1-4 (standard normal time distribution)

| | | target | | | | | | |
|---|---|---|---|---|---|---|---|---|
| | | 1-grams | 2-grams | 3-grams | 4-grams | 5-grams | 6-grams | 7-grams | total |
| source | 1-grams | **255,443** | 132,779 | 13,803 | 469 | 0 | 0 | 0 | 402,494 |
| | 2-grams | 134,458 | **217,500** | 75,441 | 8,612 | 0 | 0 | 0 | 436,011 |
| | 3-grams | 15,025 | 86,973 | **142,091** | 48,568 | 0 | 0 | 0 | 292,657 |
| | 4-grams | 635 | 10,516 | 61,741 | **98,961** | 0 | 0 | 0 | 171,853 |
| | 5-grams | 0 | 0 | 0 | 0 | 0 | 0 | 0 | 0 |
| | 6-grams | 0 | 0 | 0 | 0 | 0 | 0 | 0 | 0 |
| | 7-grams | 0 | 0 | 0 | 0 | 0 | 0 | 0 | 0 |
| | total | 405,561 | 447,768 | 293,076 | 156,610 | 0 | 0 | 0 | 1,303,015 |

Table 7. Evaluation results (MGIZA++, the original Anymalign (baseline), and Anymalign1-N).

| | BLEU | | NIST | | WER | | TER | |
|---|---|---|---|---|---|---|---|---|
| MGIZA++ | 0.2742 | | 6.6747 | | 0.5714 | | 0.6170 | |
| Anymalign | 0.2285 | | 6.0764 | | 0.6186 | | 0.6634 | |
| Anymalign1-N | equal time configuration | | | | std. norm. time distribution | | | |
| | BLEU | NIST | WER | TER | BLEU | NIST | WER | TER |
| Anymalign1-10 | 0.2182 | 5.8534 | 0.6475 | 0.6886 | 0.2361 | 6.1803 | 0.6192 | 0.6587 |
| Anymalign1-9 | 0.2296 | 6.0261 | 0.6279 | 0.6722 | 0.2402 | 6.1928 | 0.6136 | 0.6564 |
| Anymalign1-8 | 0.2253 | 5.9777 | 0.6353 | 0.6794 | 0.2366 | 6.1639 | 0.6151 | 0.6597 |
| Anymalign1-7 | 0.2371 | 6.2107 | 0.6157 | 0.6559 | 0.2405 | 6.2124 | 0.6136 | 0.6564 |
| Anymalign1-6 | 0.2349 | 6.1574 | 0.6193 | 0.6634 | 0.2403 | 6.1595 | 0.6165 | 0.6589 |
| Anymalign1-5 | 0.2376 | 6.2331 | 0.6099 | 0.6551 | 0.2436 | 6.2426 | 0.6134 | 0.6548 |
| Anymalign1-4 | **0.2423** | 6.2087 | 0.6142 | 0.6583 | **0.2442** | 6.2844 | **0.6071** | 0.6526 |
| Anymalign1-3 | 0.2403 | **6.3009** | **0.6075** | 0.6507 | 0.2441 | **6.2928** | 0.6079 | **0.6517** |
| Anymalign1-2 | 0.2406 | 6.2789 | 0.6121 | 0.6536 | 0.2404 | 6.2674 | 0.6121 | 0.6535 |
| Anymalign1-1 | 0.1984 | 5.6353 | 0.6818 | 0.7188 | 0.1984 | 5.6353 | 0.6818 | 0.7188 |

Again, we investigated the number of entries in Anymalign1-N run with this normal time distribution. We compare the number of entries in Table 6 in Anymalign1-4 with (c) equal time configuration and (d) standard normal time distribution. The number of phrase pairs on the diagonal roughly doubled when using standard normal time distribution. We can see a significant increase in the number of phrase pairs of similar lengths, while the number of phrase pairs with different lengths tends to decrease slightly. This means that the standard normal time distribution allowed us to produce much more numerous useful alignments (a priori, phrase pairs with similar lengths), while maintaining the noise (phrase pairs with different lengths) to a low level, which is a neat advantage over the original method.

## 5. MERGING PHRASE TRANSLATION TABLES

In order to check exactly how different the phrase translation table of MGIZA++ and that of Anymalign are, we performed a fourth set of experiments in which MGIZA++'s translation table is merged with that of Anymalign baseline and we used the union of the two phrase translation tables. As for feature scores in phrase translation tables for the intersection part of both aligners, i.e., entries in two translation tables share the same phrase pairs but with different feature scores, we adopted parameters computed either by MGIZA++ or by Anymalign for evaluation.

In addition, we used the feature *Backoff model* in Moses. This feature allows the use of two phrase translation tables in the process of decoding. The second phrase translation table is used as a backoff for unknown words (i.e., words that cannot be found in the first phrase translation table). To examine how good 1-grams are produced by Anymalign and how they can benefit a machine translation system, we used MGIZA++ as the first table and Anymalign baseline as the backoff table for unknown words in the experiments. We also experimented on limiting the n-grams that were used from backoff table.

Evaluation results on machine translation tasks with merged translation tables are given in Table 8.This setting outperforms MGIZA++ on BLEU scores, as well as three other evaluation metrics. The phrase translation table with Anymalign parameters for the intersection part is slightly behind the phrase translation table with MGIZA++ parameters. This may indicate that the feature scores in Anymalign phrase translation table need to be revised. In Anymalign, the frequency counts of phrase pairs are collected from subcorpora. A possible revision of computation of feature scores would be to count the number of phrase pairs from the whole corpus.

Table 8. Evaluation results (MGIZA++, the original Anymalign (baseline), merged translation tables, and backoff models).

|  | BLEU | NIST | WER | TER |
|---|---|---|---|---|
| MGIZA++ | 0.2742 | 6.6747 | 0.5714 | 0.6170 |
| Anymalign | 0.2285 | 6.0764 | 0.6186 | 0.6634 |
| Merge (Anymalign param.) | **0.2747** | **6.7101** | **0.5671** | **0.6128** |
| Merge (MGIZA++ param.) | **0.2754** | **6.7060** | **0.5685** | **0.6142** |
| Backoff model (1-grams) | **0.2809** | **6.7546** | **0.5634** | **0.6080** |
| Backoff model (2-grams) | 0.2809 | 6.7546 | 0.5634 | 0.6080 |
| Backoff model (3-grams) | 0.2804 | 6.7546 | 0.5634 | 0.6081 |
| Backoff model (4-grams) | 0.2805 | 6.7547 | 0.5634 | 0.6082 |
| Backoff model (5-grams) | 0.2804 | 6.7546 | 0.5633 | 0.6081 |
| Backoff model (6-grams) | 0.2804 | 6.7546 | 0.5633 | 0.6081 |
| Backoff model (7-grams) | 0.2804 | 6.7546 | 0.5633 | 0.6081 |

Evaluation results on using backoff models show that unigrams produced by Anymalign help in reducing the number of unknown words and thus contribute to the increase in BLEU scores. To analyze furthermore, the number of unique n-grams in the test set (French) that can be found in phrase translation tables is shown in Table 9. Anymalign gives greater lexical (1-grams) coverage than MGIZA++ and it reduces the number of unknown words on the test corpus. There are 341 unique unigrams from the French test corpus that cannot be found in the MGIZA++'s phrase translation table. These unigrams are unknown words for the MGIZA++ table and they are about twice the number of unknown words for Anymalign phrase translation table. For n-grams (n ≥ 2), Anymalign gives less coverage than MGIZA++. An analysis of overlaps and differences between two phrase translation tables is given in Table 10. It shows that 7% of phrase pairs produced by Anymalign overlap with those of MGIZA++. This shows clearly that the two methods produce different phrases.

Table 9. Number of unique n-grams in French test set found in phrase translation tables.

| n-grams | corpus | MGIZA++ | | Anymalign | |
|---|---|---|---|---|---|
|  |  | in TT | not in TT | in TT | not in TT |
| 1-gram | 3885 | **3544** | **341** | **3696** | **189** |
| 2-gram | 15230 | 10492 | 4738 | 4959 | 10271 |
| 3-gram | 22777 | 8987 | 13790 | 1179 | 21598 |
| 4-gram | 25024 | 4418 | 20606 | 212 | 24812 |
| 5-gram | 25184 | 1748 | 23436 | 46 | 25138 |
| 6-gram | 24666 | 672 | 23994 | 12 | 24654 |
| 7-gram | 23880 | 271 | 23609 | 5 | 23875 |

Table 10. Analysis of overlap between two phrase translation tables.

| Aligner | Overlap | Difference | Total |
|---|---|---|---|
| MGIZA++ | 90,086 | 3,678,331 | 3,768,417 |
| Anymalign | 90,086 | 1,281,779 | 1,371,865 |

# 6. CONCLUSIONS AND FUTURE WORK

In this section, we summarize the work of this research and highlight its contributions. In addition, we suggest possible directions for future work.

In this paper, by examining the distribution of n-grams in Anymalign phrase translation tables, we presented a method to improve the translation quality of sampling-based sub-sentential alignment approach implemented in Anymalign: firstly, Anymalign was forced to align n-grams as if they were unigrams; secondly, time was unevenly distributed over subtables; thirdly, merging of two aligners' phrase translation tables was introduced. A baseline statistical machine translation system was built to compare the translation performance of two aligners: MGIZA++ and Anymalign. Anymalign1-N, the method presented here, obtains significantly better results than the original method, the best performance being obtained with Anymalign1-4. Merging Anymalign's phrase translation table with that of MGIZA++ allows outperforming MGIZA++ alone. The use of backoff models shows that Anymalign is good for reducing the number of unknown words.

There are arguments as from which phrase length the translation quality would benefit. [13] suggested that phrase length up to three words contributes the most to BLEU scores, which was confirmed for instance by [25, 26]. However, [27] argued that longer phrases should not be neglected. As for Anymalign1-N, Anymalign1-3 and Anymalign1-4, in which phrase lengths are limited to three and four words respectively, get the best results in four evaluation metrics among all variants of Anymalign. This would confirm that longer phrases could indeed be a source of noise in translation process. On the other hand, more reliable, shorter phrases, would contribute a lot to translation quality.

In addition, a recent work by [28] shows that it is safe to discard a phrase if it can be decomposed in shorter phrases. They note that discarding the phrase *the French government*, which is compositional, does not change translation cost. On the other hand, the phrase *the government of France* should be retained in the phrase translation table. It would raise the question on what phrases constitute a *good* phrase translation table. As for Anymalign, we might observe the proportion of compositional and non-compositional phrases in its phrase translation table for future work.

Furthermore, according to the differences of the evaluation results between using Anymalign feature scores and those of MGIZA++ in the overlapping part of their respective phrase translation tables, we wonder whether the feature scores computed by Anymalign should be modified in order to mimic those of MGIZA++ and better suit the expectation of the Moses decoder. Also there is the concern on whether the distribution of phrase pairs in MGIZA++'s translation table is ideal and what possibly different distribution in Anymalign's translation tables would contribute to further improvement. This is an important aspect for further research.

## ACKNOWLEDGEMENTS

Part of the research presented in this paper has been done under a Japanese grant-in-aid (Kakenhi C, 23500187: Improvement of alignments and release of multilingual syntactic patterns for statistical and example-based machine translation).

## REFERENCES

[1]   F.J. Och and H. Ney, "A systematic comparison of various statistical alignment models," Computational Linguistics, vol.29, no.1, pp.19-51, 2003.

[2]   P. Brown, S. Della Pietra, V. Della Pietra, and R. Mercer, "The mathematics of statistical machine translation: Parameter estimation," Computational Linguistics, vol.19, no.2, pp.263-311, 1993.

[3]   S. Vogel, H. Ney, and C. Tillman, "Hmm-based word alignment in statistical translation," Proceedings of 16th International Conference on Computational Linguistics, Copenhagen, pp.836-841, 1996.

[4]   Q. Gao and S. Vogel, "Parallel implementations of word alignment tool," Software Engineering, Testing, and Quality Assurance for Natural Language Processing, Columbus, Ohio, pp.49-57, 2008.

[5]   A. Lardilleux and Y. Lepage, "Sampling-based multilingual alignment," Proceedings of International Conference on Recent Advances in Natural Language Processing, Borovets, Bulgaria, pp.214-218, 2009.

[6]   A. Lardilleux, Y. Lepage, and F. Yvon, "The contribution of low frequencies to multilingual sub-sentential

alignment: a differential associative approach," International Journal of Advanced Intelligence, vol.3, no.2, pp.189-217, 2011.

[7]   W. Gale and K. Church, "Identifying word correspondences in parallel texts," Proceedings of 4th DARPA workshop on Speech and Natural Language, Pacific Grove, pp.152-157, 1991.

[8]   D. Melamed, "Models of translational equivalence among words," Computational Linguistics, vol.26, no.2, pp.221-249, 2000.

[9]   R. Moore, "Association-based bilingual word alignment," Proceedings of ACL Workshop on Building and Using Parallel Texts, Ann Arbor, pp.1-8, 2005.

[10]  P. Liang, B. Taskar, and D. Klein, "Alignment by agreement," Proceedings of Human Language Technology Conference of the NAACL, New York City, pp.104-111, 2006.

[11]  A. Toral, M. Poch, P. Pecina, and G. Thurmair, "Efficiency-based evaluation of aligners for industrial applications," Proceedings of 16th Annual Conference of the European Association for Machine Translation, pp.57-60, 2012.

[12]  A. Lardilleux and Y. Lepage, "Hapax Legomena : their contribution in number and efficiency to word alignment," Lecture notes in computer science, vol.5603, pp.440-450, 2009.

[13]  P. Koehn, F.J. Och, and D. Marcu, "Statistical phrase-based translation," Proceedings of 2003 Human Language Technology Conference of the North American Chapter of the Association for Computational Linguistics, Edmonton, pp.48-54, 2003.

[14]  P. Koehn, "Europarl: A Parallel Corpus for Statistical Machine Translation," Proceedings of 10th Machine Translation Summit (MT Summit X), Phuket, pp.79-86, 2005.

[15]  P. Koehn, H. Hoang, A. Birch, C. Callison-Burch, M. Federico, N. Bertoldi, B. Cowan,W. Shen, C. Moran, R. Zens, C. Dyer, O. Bojar, A. Constantin, and E. Herbst, "Moses: Open source toolkit for statistical machine translation," Proceedings of 45th Annual Meeting of the Association for Computational Linguistics, Prague, Czech Republic, pp.177-180, 2007.

[16]  F.J. Och, "Minimum error rate training in statistical machine translation," Proceedings of 41st Annual Meeting on Association for Computational Linguistics, Sapporo, Japan, pp.160-167, 2003.

[17]  A. Stolcke, "SRILM-an extensible language modeling toolkit," Proceedings of 7th International Conference on Spoken Language Processing, Denver, Colorado, pp.901-904, 2002.

[18]  S. Nießen, F.J. Och, G. Leusch, and H. Ney, "An evaluation tool for machine translation: Fast evaluation for machine translation research," Proceedings of 2nd International Conference on Language Resources and Evaluation, Athens, pp.39-45, 2000.

[19]  K. Papineni, S. Roukos, T. Ward, and W.J. Zhu, "BLEU: a method for automatic evaluation of machine translation," Proceedings of 40th Annual Meeting of the Association for Computational Linguistics, Philadelphia, pp.311-318, 2002.

[20]  G. Doddington, "Automatic evaluation of machine translation quality using N-gram co-occurrence statistics," Proceedings of 2nd International Conference on Human Language Technology Research, San Diego, pp.138-145, 2002.

[21]  M. Snover, B. Dorr, R. Schwartz, L. Micciulla, and J. Makhoul, "A study of translation edit rate with targeted human annotation," Proceedings of Association for Machine Translation in the Americas, Cambridge, Massachusetts, pp.223-231, 2006.

[22]  A. Lardilleux, J. Chevelu, Y. Lepage, G. Putois, and J. Gosme, "Lexicons or phrase tables? An investigation in sampling-based multilingual alignment," Proceedings of 3rd workshop on example-based machine translation, Dublin, Ireland, pp.45-52, 2009.

[23]  Y. Ma, N. Stroppa, and A. Way, "Bootstrapping word alignment via word packing," Proceedings of 45th Annual Meeting of the Association of Computational Linguistics, Prague, Czech Republic, pp.304-311, 2007.

[24]  A.C. Henríquez Q., R.M. Costa-jussà, V. Daudaravicius, E.R. Banchs, and B.J. Mariño, "Using collocation segmentation to augment the phrase table," Proceedings of Joint FifthWorkshop on Statistical Machine Translation and MetricsMATR, Uppsala, Sweden, pp.98-102, 2010.

[25]  Y. Chen, M. Kay, and A. Eisele, "Intersecting multilingual data for faster and better statistical translations," Proceedings of Human Language Technologies: The 2009 Annual Conference of the North American Chapter of the Association for Computational Linguistics, Boulder, Colorado, pp.128-136, 2009.

[26]  G. Neubig, T. Watanabe, E. Sumita, S. Mori, and T. Kawahara, "An unsupervised model for joint phrase alignment and extraction," Proceedings of 49th Annual Meeting of the Association for Computational Linguistics: Human Language Technologies, Portland, Oregon, USA, pp.632-641, 2011.

[27]  C. Callison-Burch, C. Bannard, and J. Schroeder, "Scaling phrase based statistical machine translation to larger corpora and longer phrases," Proceedings of 43rd Annual Meeting on Association for Computational Linguistics, Ann Arbor, Michigan, pp.255-262, 2005.

[28]  R. Zens, D. Stanton, and P. Xu, "A systematic comparison of phrase table pruning technique," Proceedings of Joint Conference on Empirical Methods in Natural Language Processing and Computational Natural Language Learning, Jeju Island, Korea, pp.972-983, 2012.